\documentclass[10pt,twocolumn,letterpaper]{article}

\usepackage{iccv}
\usepackage{times}
\usepackage{bm}
\usepackage{epsfig}
\usepackage{graphicx}
\usepackage{enumitem}
\usepackage{microtype}
\usepackage{amsmath}
\usepackage{amssymb}
\usepackage{color}
\usepackage{caption}

\newenvironment{tight_itemize}{
\begin{itemize}[leftmargin=20pt]
\setlength{\topsep}{0pt}
\setlength{\itemsep}{0pt}
\setlength{\parskip}{0pt}
\setlength{\parsep}{0pt}
}{\end{itemize}}

\usepackage[pagebackref=true,breaklinks=true,letterpaper=true,colorlinks,bookmarks=false]{hyperref}

\iccvfinalcopy

\def\iccvPaperID{1987}

\ificcvfinal\fi
\begin{document}

\title{Towards Large-Pose Face Frontalization in the Wild}

\author{Xi Yin$^{\dagger}$\thanks{This work was supported by a research gift from NEC Labs to Michigan State University.}, Xiang Yu$^{\ddag}$, Kihyuk Sohn$^{\ddag}$, Xiaoming Liu$^{\dagger}$ and Manmohan Chandraker$^{\S\ddag}$\\
$^{\dagger}$Michigan State University \\
$^{\S}$University of California, San Diego \\
$^{\ddag}$ NEC Laboratories America\\
{\tt\small \{yinxi1,liuxm\}@cse.msu.edu, \{xiangyu,ksohn,manu\}@nec-labs.com}}

\maketitle

\begin{abstract}
Despite recent advances in face recognition using deep learning, severe accuracy drops are observed for large pose variations in unconstrained environments. Learning pose-invariant features is one solution, but needs expensively labeled large-scale data and carefully designed feature learning algorithms. In this work, we focus on frontalizing faces in the wild under various head poses, including extreme profile views. We propose a novel deep 3D Morphable Model (3DMM) conditioned Face Frontalization Generative Adversarial Network (GAN), termed as FF-GAN, to generate neutral head pose face images. Our framework differs from both traditional GANs and 3DMM based modeling. Incorporating 3DMM into the GAN structure provides shape and appearance priors for fast convergence with less training data, while also supporting end-to-end training. The 3DMM-conditioned GAN employs not only the discriminator and generator loss but also a new masked symmetry loss to retain visual quality under occlusions, besides an identity loss to recover high frequency information. Experiments on face recognition, landmark localization and 3D reconstruction consistently show the advantage of our frontalization method on faces in the wild datasets.\footnote{Detail results and resources can be refered to: \url{http://cvlab.cse.msu.edu/project-face-frontalization.html}.}
\end{abstract}

\section{Introduction}
\vspace{-2mm}
Frontalization of face images observed from extreme viewpoints is a problem of fundamental interest in both human and machine facial processing and recognition. Indeed, while humans are innately skilled at face recognition, newborns do not perform better than chance on recognition from profile views \cite{Turati_etal_2008}, although this ability seems to develop rapidly within a few months of birth \cite{Fagan_1976}. Similarly, dealing with profile views of faces remains an enduring challenge for computer vision too. The advent of deep convolutional neural networks (CNNs) has led to large strides in face recognition, with verification accuracy surpassing human levels \cite{SunCVPR14,Sunnips2014} on datasets such as LFW. But even representations learned with state-of-the-art CNNs suffer for profile views, with severe accuracy drops reported in recent studies \cite{Kanmeina16,CFP}. Besides aiding recognition, frontalization of faces is also a problem of independent interest, with potential applications such as face editing, accessorizing and creation of models for use in virtual and augmented reality.

\begin{figure}[t!]
\begin{center}
\includegraphics[width=0.45\textwidth]{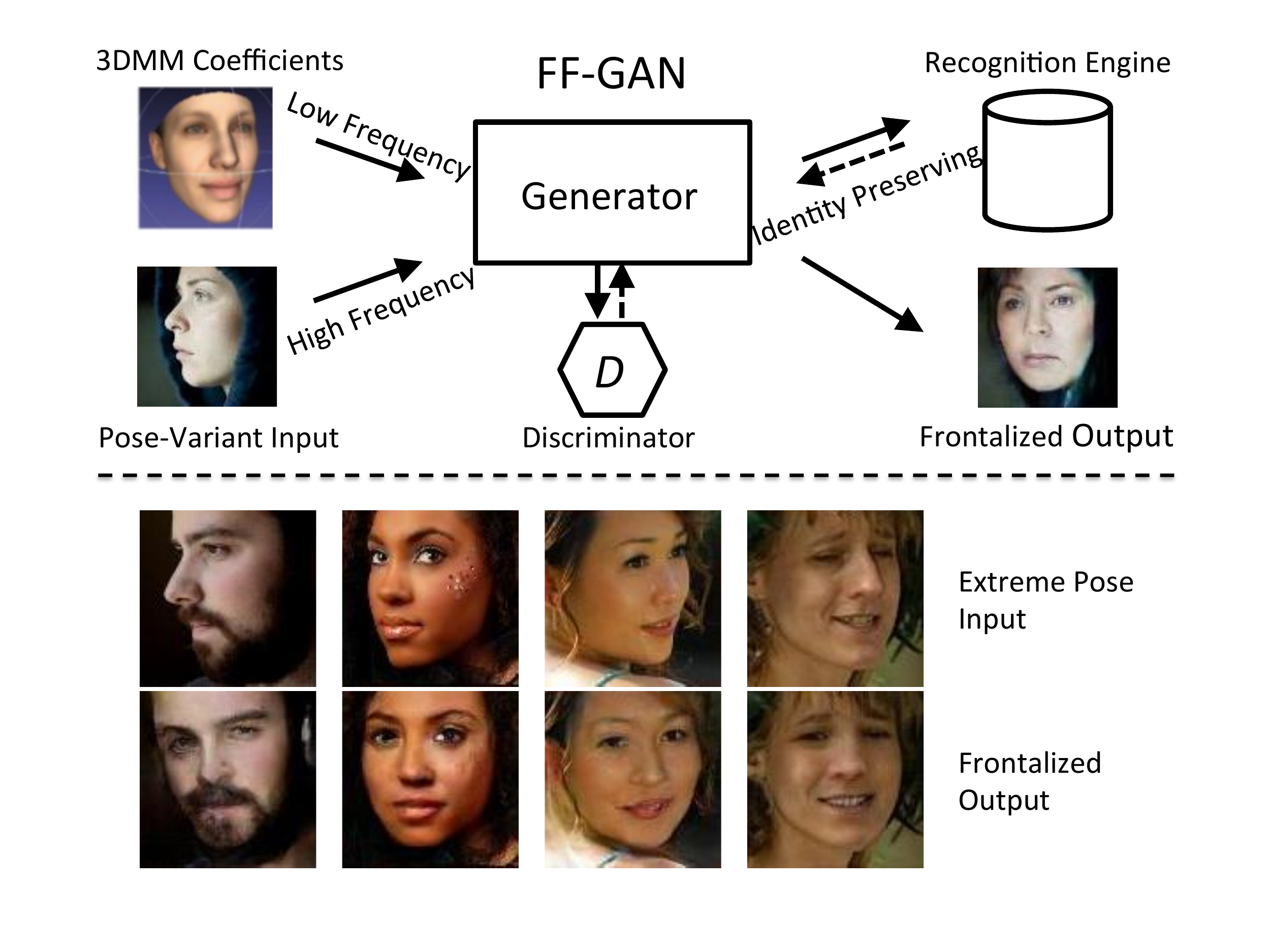}
\end{center}
\vspace{-6mm}
\caption{The proposed FF-GAN framework. Given a non-frontal face image as input, the generator produces a high-quality frontal face. Learned 3DMM coefficients provide global pose and low frequency information, while the input image injects high frequency local information. A discriminator distinguishes generated faces against real ones, where high-quality frontal faces are considered real. A face recognition engine is used to preserve identity information. The output is a high quality frontal face that retains identity.}
\vspace{-4mm}
\label{fig:concept}
\end{figure}

In recent years, CNNs have leveraged the availability of large-scale face datasets to also learn pose-invariant representations for recognition, using either a joint single framework \cite{MasiCVPR16} or multi-view pose-specific networks \cite{ZhuNIPS2014,Kanmeina16}. Early works on face frontalization in computer vision rely on frameworks inspired by computer graphics. The well-known 3D Morphable Model (3DMM) \cite{3dmm} explicitly models facial shape and appearance to match an input image as closely as possible. Subsequently, the recovered shape and appearance can be used to generate a face image under novel viewpoints. Many 3D face reconstruction methods \cite{roth2015,zhu2016face,rothpami} build upon this direction by improving speed or accuracy. Deep learning has made inroads into data-driven estimation of 3DMM models too \cite{ZhuNIPS2014,Kanmeina16,jourabloo2017pose}, circumventing some drawbacks of early methods such as over-reliance on the accuracy of 2D landmark localization~\cite{xiang_pami_cdm_2015,xiang_cor_2014}. Due to restricted Gaussian assumptions and nature of losses used, insufficient representation ability for facial appearance prevents such deep models from producing outputs of high quality. While inpainting methods such as \cite{ZhuXYCVPR15} attempt to minimize the impact on quality due to self-occlusions, they still do not retain identity information.

In this paper, we propose a novel generative adversarial network framework FF-GAN that incorporates elements from both deep 3DMM and face recognition CNNs to achieve high-quality and identity-preserving frontalization, using a single input image that can be a profile view up to $90^{\circ}$. Our framework targets more on face-in-the-wild conditions which show more challenge on illumination, head pose variation, self-occlusion and so on. To the best of our knowledge, this is the first face frontalization work that handles pose variations at such extreme poses.

Noting the challenge of purely data-driven face frontalization, Section \ref{sec:reconstruction} proposes to enhance the input to the GAN with 3DMM coefficients estimated by a deep network. This provides a useful prior to regularize the frontalization, however, it is well known that deep 3DMM reconstructions are limited in their ability to retain high-frequency information. Thus, Section \ref{sec:generation} proposes a method to combine 3DMM coefficients with the input image to generate an image that maintains both global pose accuracy and retains local information present in the input image. In particular, the generator in our GAN produces a frontal image based on a reconstruction loss, a smoothness loss and a novel symmetry-enforcing loss. The aim of the generator is to fool the discriminator, presented in Section \ref{sec:discrimination}, into being unable to distinguish the generated frontal image from a real one. However, neither the 3DMM that loses high-frequency information, nor the GAN that only aligns domain-level distributions, suffice to preserve identity information in the generated image. To retain identity information, Section \ref{sec:recognition} proposes to use a recognition engine to align the feature representations of the generated image with the input. A balanced training with all the above objectives results in high-quality frontalized faces that preserve identity.

We extensively evaluate our framework on several well-known datasets including Multi-PIE, AFLW, LFW, and IJB-A. In particular, Section \ref{sec:experiments} demonstrates that the face verification accuracy on LFW dataset that uses information from our frontalized outputs exceeds previous state of the art. We observe even larger improvements on the Multi-PIE dataset especially for the viewpoints beyond $45^\circ$ and being the sole method among recent works to produce high recognition accuracies for $75^\circ$-$90^\circ$. We present ablation studies to analyze the effect of each module and several qualitative results to visualize the quality of our frontalization.

To summarize, our key contributions are:
\vspace{-0.2cm}
\begin{tight_itemize}
\item A novel GAN-based end-to-end deep framework to achieve face frontalization even for extreme viewpoints.
\item A deep 3DMM model to provide shape and appearance regularization beyond the training data.
\item Effective symmetry-based loss and smoothness regularization that lead to generation of high-quality images.
\item Use of a deep face recognition CNN to enforce that generated faces satisfy identity-preservation, besides realism and frontalization.
\item Consistent improvements on several datasets across multiple tasks, such as face recognition, landmark localization and 3D reconstruction.
\end{tight_itemize}

\begin{figure*}[t!]
\begin{center}
\includegraphics[width=0.85\textwidth]{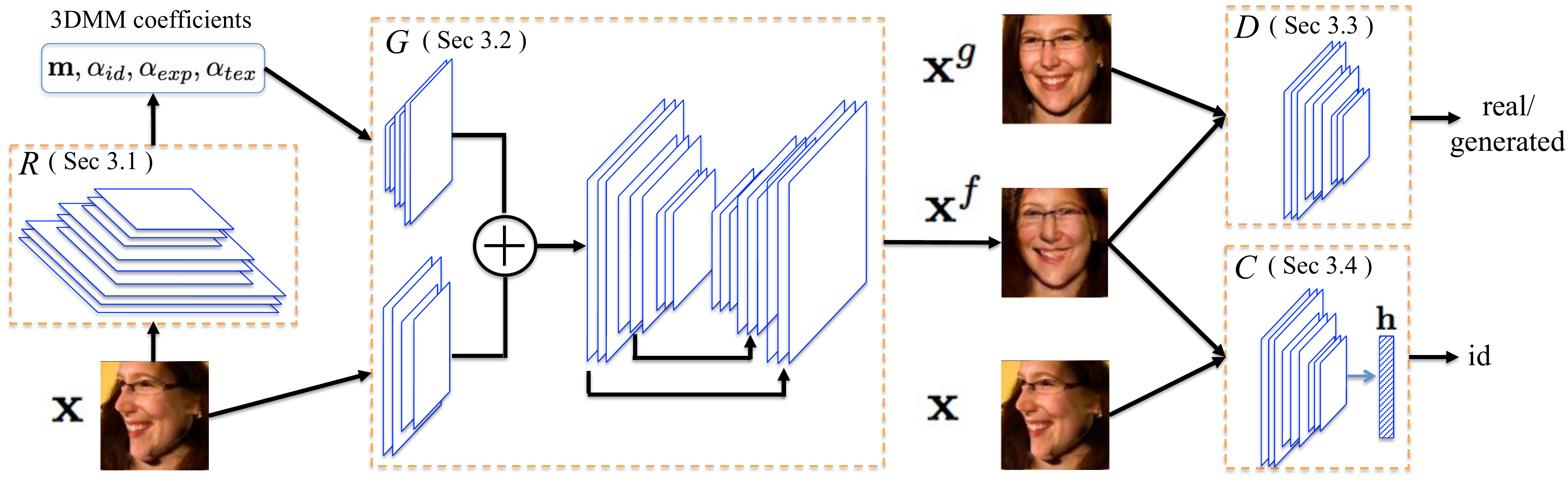}
\end{center}
\vspace{-6mm}
\caption{The proposed FF-GAN for large-pose face frontalization. 
$R$ is the reconstruction module for 3DMM coefficients estimation. 
$G$ is the generation module to synthesize a frontal face.
$D$ is the discrimination module to make real or generated decision. 
$C$ is the recognition module for identity classification. }
\vspace{-4mm}
\label{fig:overview}
\end{figure*}

\section{Related Work}
\vspace{-0.2cm}
\paragraph{Face Frontalization}
Synthesizing a frontal view of a face from a single image with large pose variation is very challenging because recovering the 3D information from 2D projections is ambiguous and there exists self-occlusion. A straightforward method is to build 3D models for faces and directly rotate the 3D face models. Seminal works date back to the 3D Morphable Model (3DMM)~\cite{3dmm}, which models both the shape and appearance as PCA spaces. Such 3DMM fitting helps boost 3D face reconstruction \cite{roth2015} and 3D landmark localization \cite{jourabloo2015pose,zhu2016face} performance. Hassner et al.~\cite{hassner2014} apply a shared 3D shape model combined with input images to produce the frontalized appearance. Ferrari et al.~\cite{ferrari2016} use 3DMM to fit to the input image and search for the dense point correspondence to complete the occlusion region. Zhu et al.~\cite{ZhuXYCVPR15} provide a high fidelity pose and expression normalization method based on 3DMM. 

Besides model-based methods, Sagonas et al.~\cite{sagonas2015} propose frontalization as a low-rank constraint optimization problem to target landmark localization. Some deep learning based methods also show promising performance in terms of pose rectification. In~\cite{yang2015}, a recurrent transform unit is proposed to synthesize discrete 3D views. A concatenated network structure is applied in \cite{yim2015} to rotate the face, where the output is regularized by the image level reconstruction loss. In~\cite{forrester2017,bousmalis2017}, a perception loss is designed to supervise the generator training. Our method is also based on deep learning. We incorporate 3DMM into a deep network and propose a novel GAN based framework to jointly utilize the geometric prior from 3DMM and high frequency information from the original image. A discriminator and a face recognition engine are proposed to regularize the generator to preserve identity.

\vspace{-0.6cm}
\paragraph{Generative Adversarial Networks} Introduced by Goodfellow et al.~\cite{goodfellow2014generative}, the Generative Adversarial Network (GAN) maps from a source data distribution to a target distribution using a minimax optimization over a generator and a discriminator. GANs have been used for a wide range of applications including image generation~\cite{denton2015,brox2015,taigman2017}, 3D object generation~\cite{3dgan}, etc. Deep Convolutional GAN (DC-GAN)~\cite{dcgan} extended the original GAN multi-layer perceptron network into convolutional structures. Many methods favor conditional settings by introducing latent factors to disentangle the objective space and thereby achieve better synthesis. For instance, Info-GAN~\cite{infogan} employs the latent code for information loss to regularize the generative network. 

A recent method, DR-GAN \cite{tran2017}, has been proposed concurrent to this work. It also uses a recognition engine to regularize the identity, while using a pose code as input to the encoder to generate an image with specific pose. Instead of explicitly injecting the latent code, our framework learns the shape and appearance code from a differentiable 3DMM deep network, which supports the flavor of end-to-end joint optimization for the GAN. Unlike our framework, it is the encoder in \cite{tran2017} that is required to be identity-preserving, which suffices for reconstruction, but results in loss of spatial and high-frequency information crucial for image generation.

\vspace{-0.6cm}
\paragraph{Pose-Invariant Feature Representation}
While face frontalization may be considered an image-level pose-invariant representation, feature representations invariant to pose have also been a mainstay of face recognition. Early works employ Canonical Correlation Analysis (CCA) to analyze the commonality among pose-variant samples. Recently, deep learning based methods consider several aspects, such as multiview perception layers~\cite{ZhuNIPS2014}, to learn a model separating identity from viewpoints. Feature pooling across different poses~\cite{Kanmeina16} is also proposed to allow a single network structure for multiple pose inputs. Pose-invariant feature disentanglement \cite{xipeng2017} or identity preservation \cite{zhu2013deep,xiyin2017} methods aim to factorize out the non-identity part with carefully designed networks. Some other methods focus on fusing information at the feature level \cite{chen2016unconstrained} or distance level \cite{MasiCVPR16}. Our method is mostly related to identity preserving methods, in which we apply a face recognition engine as an additional discriminator to guide the generator for better synthesis, while it updates itself towards better discriminative capability for identity classification.

\vspace{-2mm}
\section{Proposed Approach}
\vspace{-2mm}
\label{sec:method}
The mainstay of FF-GAN is a generative adversarial network that consists of a generator $G$ and a discriminator $D$. $G$ takes a non-frontal face as input to generate a frontal output, while $D$ attempts to classify it as a real frontal image or a generated one. Additionally, we include a face recognition engine $C$ that regularizes the generator output to preserve identity features. A key component is a deep 3DMM model $R$ that provides shape and appearance priors to the GAN that play a crucial role in alleviating the difficulty of large pose face frontalization. Figure \ref{fig:overview} summarizes our framework.

Let $\mathbb{D} =\{ {{\bf{x}}_i, {\bf{x}}_i^g, {\bf{p}}_i^g}, y_i\}_{i=1}^{N}$ be the training set with $N$ samples, with each sample consisting of an input image ${\bf{x}}_i$ with arbitrary pose, a corresponding ground truth frontal face ${\bf{x}}_i^g$, the ground truth 3DMM coefficients ${\bf{p}}_i^g$ and the identity label $y_i$. We henceforth omit the sample index $i$ for clarity. 

\subsection{Reconstruction Module}
\vspace{-2mm}
\label{sec:reconstruction}
Frontalization from extreme pose variation is a challenging problem. While a purely data-driven approach might be possible given sufficient data and an appropriate training regimen, however it is non-trivial. Therefore, we propose to impose a prior on the generation process, in the form of a 3D Morphable Model (3DMM) \cite{3dmm}. This reduces the training complexity and leads to better empirical performance with limited data.

Recall that 3DMM represents faces in the PCA space:
\vspace{-2mm}
\begin{equation}
\begin{aligned}
{\bf{S}} {} & = {\bf{ \bar{S}}} + {\bf{A}}_{id}{\bf{\alpha}}_{id} + {\bf{A}}_{exp}{\bf{\alpha}}_{exp}, \\
{\bf{T}} {} & = {\bf{\bar{T}}} + {\bf{A}}_{tex}{\bf{\alpha}}_{tex},
\label{eq:3d}
\end{aligned}
\vspace{-3mm}
\end{equation}
where ${\bf{S}}$ are the 3D shape coordinates computed as the linear combination of the mean shape $\bf{\bar{S}}$, the shape basis ${\bf{A}}_{id}$, and the expression basis ${\bf{A}}_{exp}$, while ${\bf{T}}$ is the texture that is the linear combination of the mean texture $\bf{\bar{T}}$ and the texture basis ${\bf{A}}_{tex}$. 
The coefficients $\{ {\bf{\alpha}}_{id}, {\bf{\alpha}}_{exp}, {\bf{\alpha}}_{tex} \}$ define a unique 3D face. 

Previous work~\cite{jourabloo2015pose, zhu2016face} applies 3DMM for face alignment where a weak perspective projection model is used to project the 3D shape into 2D space. Similar to \cite{jourabloo2015pose}, we optimize a projection matrix ${\bf{m}}\in\mathbb{R}^{2\times4}$ based on pitch, yaw, roll, scale and 2D translations to represent the pose of an input face image. Let ${\bf{p}} = \{ {\bf{m}}, {\bf{\alpha}}_{id}, {\bf{\alpha}}_{exp}, {\bf{\alpha}}_{tex}\}$ denotes the 3DMM coefficients. The target of the reconstruction module $R$ is to estimate ${\bf{p}} = R({\bf{x}})$, given an input image ${\bf{x}}$. Since the intent is for $R$ to also be trainable with the rest of the framework, we use a CNN model based on CASIA-Net~\cite{yi2014learning} for this regression task. We apply $z$-score normalization to each dimension of the parameters before training. A weighted parameter distance cost similar to \cite{zhu2016face} is used:
\vspace{-2mm}
\begin{equation}
\min_{\bf{p}}L_R = ({\bf{p}} - {\bf{p}}^g)^\top {\bf{W}} ({\bf{p}} - {\bf{p}}^g),
\label{eq:objR}
\vspace{-3mm}
\end{equation}
where ${\bf{W}}$ is the importance matrix whose diagonal is the weight of each parameter.

\subsection{Generation Module}
\vspace{-2mm}
\label{sec:generation}
The pose estimate obtained from 3DMM is quite accurate, however, frontalization using it leads to loss of high frequency details present in the original image. This is understandable, since a low-dimensional PCA representation can preserve most of the energy in lower frequency components. Thus, we use a generative model that relies on 3DMM coefficients ${\bf{p}}$ and the input image ${\bf{x}}$ to recover a frontal face that preserves both the low and high frequency components.

In Figure \ref{fig:overview}, features from the two inputs to the generator $G$ are fused through an encoder-decoder network to synthesize a frontal face ${\bf{x}}^f = G({\bf{x}}, {\bf{p}})$. To penalize the output from ground truth ${\bf{x}}^g$, the straight-forward objective is the reconstruction loss that aims at reconstructing the ground truth with minimal error:
\vspace{-3mm}
\begin{equation}
L_{G_{rec}} = \| G({\bf{x}}, {\bf{p}}) - {\bf{x}}^g\|_1.
\label{eqn:grec}
\vspace{-2mm}
\end{equation}
Since an $L_2$ loss empirically leads to blurry output, we use an $L_1$ loss to better preserve high frequency signals. At the beginning of training, the reconstruction loss harms the overall process since the generation is far from frontalized, so the reconstruction loss operates on a poor set of correspondences. Thus, the weight for reconstruction loss should be set in accordance to the training stage. The details of tuning the weight are discussed in Section~\ref{sec:implementation}.

To reduce block artifacts, we use a spatial total variation loss to encourage smoothness in the generator output:
\vspace{-3mm}
\begin{equation}
L_{G_{tv}} = \frac{1}{|\Omega|}\int_{\Omega}|\nabla G({\bf{x}}, {\bf{p}})|du,
\label{eqn:gtv}
\vspace{-2mm}
\end{equation}
where $|\nabla G|$ is the image gradient, $u\in\mathbb{R}^{2}$ is the two dimensional coordinate increment, $\Omega$ is the image region and $|\Omega|$ is the area normalization factor.

Based on the observation that human faces share self-similarity across left and right halves, we explicitly impose a symmetry loss. We recover a frontalized 2D projected silhouette, $\mathcal{M}$, from the frontalized 3DMM model indicating the visible parts of the face. The mask $\mathcal{M}$ is binary, with nonzero values indicating visible regions and zero otherwise. Similar masking constraint is shown effective in recent work~\cite{tvsn_cvpr2017}. By horizontally flipping the face, we achieve another mask $\mathcal{M}_{flip}$. We demand that generated frontal images for the original input image and its flipped version should be similar within their respective masks:
\vspace{-3mm}
\begin{align}
&L_{G_{sym}} = \|\mathcal{M}\odot G({\bf{x}},{\bf{p}}) - \mathcal{M}\odot G({\bf{x}}_{flip}, {\bf{p}}_{flip})\|_{2} \notag \\
&+ \|{\mathcal{M}}_{flip}\odot G({\bf{x}},{\bf{p}}) - {\mathcal{M}}_{flip}\odot G({\bf{x}}_{flip}, {\bf{p}}_{flip})\|_{2}.
\label{eqn:gtv}
\vspace{-2mm}
\end{align}
Here, ${\bf{x}}_{flip}$ is the horizontal flipped image for input $\bf{x}$, ${\bf{p}}_{flip}$ are the 3DMM coefficients for ${\bf{x}}_{flip}$ and $\odot$ denotes element-wise multiplication. Note that the role of the mask is to focus on the visible face portion, rather than invisible face portion and background.

\subsection{Discrimination Module}
\vspace{-2mm}
\label{sec:discrimination}
Generative Adversarial Networks~\cite{goodfellow2014generative}, formulated as a two-player game between a generator and a discriminator, have been widely used for image generation. In this work, the generator $G$ synthesizes a frontal face image ${\bf{x}}^f$ and the discriminator $D$ distinguishes between the generated face from the frontal face ${\bf{x}}^g$. Note that in a conventional GAN, all images used for training are considered as real samples. However, we limit ``real'' faces to frontal views only. Thus, $G$ must generate both realistic and frontal face images. 

Our $D$ consists of five convolutional layers and one linear layer that generates a 2D vector with each dimension representing the probability of the input to be real or generated. During training, $D$ is updated with two batches of samples in each iteration. The following objective is minimized: 
\vspace{-2mm}
\begin{equation}
\small
\min_{D} L_D = -\mathbb{E}_{{\bf{x}}^g\in \mathcal{R}}\log(D({\bf{x}}^g)) - \mathbb{E}_{{\bf{x}}\in \mathcal{K}}\log(1-D(G({\bf{x}}, {\bf{p}}))),
\label{eq:objD}
\vspace{-1mm}
\end{equation}
where $\mathcal{R}$ and $\mathcal{K}$ are the real and generated image sets respectively.

On the other hand, $G$ aims to fool $D$ to classify the generated image to be real with the following loss:
\vspace{-2mm}
\begin{align}
L_{G_{gan}} = -\mathbb{E}_{{\bf{x}}\in \mathcal{K}}\log(D(G({\bf{x}},{\bf{p}}))).
\vspace{-3mm}
\end{align}

The competition between $G$ and $D$ improves both modules. In the early stages when face images are not fully frontalized, $D$ focuses on the pose of the face to make the real or generated decision, which in turn helps $G$ to generate a frontal face. In the later stages when face images are frontalized, $D$ focuses on subtle details of frontal faces, which guides $G$ to generate a realistic frontal face that is difficult to achieve with the supervisions of \eqref{eqn:grec} and \eqref{eqn:gtv} alone. 

\subsection{Recognition Module}
\vspace{-2mm}
\label{sec:recognition}
A key challenge in large-pose face frontalization is to preserve the original identity. This is a difficult task due to self-occlusion in profile faces. The discriminator above can only determine whether the generated image is realistic and in frontal pose, but does not consider whether identity of the input image is retained. Although we have L1, total variation and masked symmetry losses for face generation, they treat each pixel equally that result in the loss of discriminative power for identity features. Therefore, we use a recognition module $C$ to impart correct identity to the generated images. 

We use a CASIA-Net structure for the recognition engine $C$, with a cross-entropy loss for training $C$ to classify image ${\bf{x}}$ with the ground truth identity $y$:
\vspace{-2mm}
\begin{equation}
\small
\min_{C}L_C = \sum_{j}-y_j\log(C_j({\bf{x}})),
\label{eq:objC}
\vspace{-2mm}
\end{equation}
where $j$ is the index of the identity classes. Now, our generator $G$ is regularized by the signal from $C$ to preserve the same identity as the input image. If the identity label of the input image is not available, we regularize the extracted identity features ${\bf{h}}^f$ of the generated image to be similar to those of the input image, denoted as ${\bf{h}}$. During training, $C$ is updated with real input images to retain discriminative power and the loss from the generated images is back-propagated to update the generator $G$:
\vspace{-2mm}
\begin{equation}
\small
L_{G_{id}} =
\begin{cases}
 \sum_{j}-y_j\log(C_j(G({\bf{x}}, {\bf{p}}))) , & \exists y \\
\| {\bf{h}}^f - {\bf{h}}\|_2^2, & \nexists y.
\end{cases}
\vspace{-2mm}
\end{equation}

To summarize the framework, the reconstruction module $R$ provides guidance to the frontalization through \eqref{eq:objR}, the discriminator does so through \eqref{eq:objD} and the recognition engine through \eqref{eq:objC}. Thus, the generator combines all these sources of information to optimize an overall objective function:
\vspace{-2mm}
\begin{align}
\small
&\min_{G} L_{G}=\lambda_{rec}L_{G_{rec}} + \lambda_{tv}L_{G_{tv}} \notag \\
&+ \lambda_{sym}L_{G_{sym}} + \lambda_{gan}L_{G_{gan}} + \lambda_{id}L_{G_{id}}.
\label{eq:objall}
\vspace{-2mm}
\end{align}
The weights above are discussed in Section~\ref{sec:implementation} to illustrate how each component contributes to joint optimization of $G$.

\vspace{-1mm} 
\section{Implementation Details}
\label{sec:implementation}
\vspace{-2mm}
Our framework consists of mainly four parts as shown in Figure~\ref{fig:overview}, the deep 3DMM reconstructor $R$, a two-way fused encoder-decoder generator, the real/generated discriminator and a face recognition engine jointly trained for the identity regularization. All the detailed network structures are introduced in the appendix. It is difficult to initialize the overall network from scratch. The generator $G$ is expected to receive the correct 3DMM coefficients, whereas the reconstructor needs to be pre-trained. Our identity regularization also requires correct identity information from the recognizer. Thus, the reconstructor $R$ is pre-trained until we achieve comparable performance for face alignment compared to previous work~\cite{zhu2016face} using 300W-LP~\cite{zhu2016face}. The recognizer is pre-trained using CASIA-WebFace~\cite{yi2014learning} to achieve good verification accuracy on LFW.

The end-to-end joint training is conducted after $R$ and $C$ are ready. Notice that we leave generator $G$ and discriminator $D$ training from scratch simultaneously because we believe pre-trained $G$ and $D$ do not contribute much to the adversarial training. Good $G$ with poor $D$ will quickly pull $G$ to be poor again and vice versa. Further these two components should also match with each other. Good $G$ may be evaluated poor by a good $D$ as the discriminator may be trained from other sources. As shown in Equation~\ref{eq:objall}, we set up five balance factors to control the loss contribution to the overall loss. The end-to-end training can be divided into three steps. For the first step, $\lambda_{rec}$ is set to $0$ and $\lambda_{id}$ is set to 0.01, since these two parts are highly related with the mapping from the generated output to the reference input. Typical values for $\lambda_{tv}$, $\lambda_{sym}$ and $\lambda_{gan}$ are all 1.0. Once the training error of $G$ and $D$ strikes a balance within usually 20 epochs, we change $\lambda_{rec}$ and $\lambda_{id}$ to be 1.0 while tuning down $\lambda_{tv}$ to be 0.5, $\lambda_{sym}$ to be 0.8, respectively. It takes another 20 epochs to strike a new balance. Note that we fix model $C$ for such two-stage training. After that, we relax model $C$ and further fine-tune all the modules jointly. Further details are included in the supplementary materials. 
Network structures and more results are refered to: \url{http://cvlab.cse.msu.edu/project-face-frontalization.html}.

\section{Experiments}
\label{sec:experiments}
\vspace{-2mm}

\begin{figure*}
\begin{center}
\includegraphics[width=0.85\textwidth]{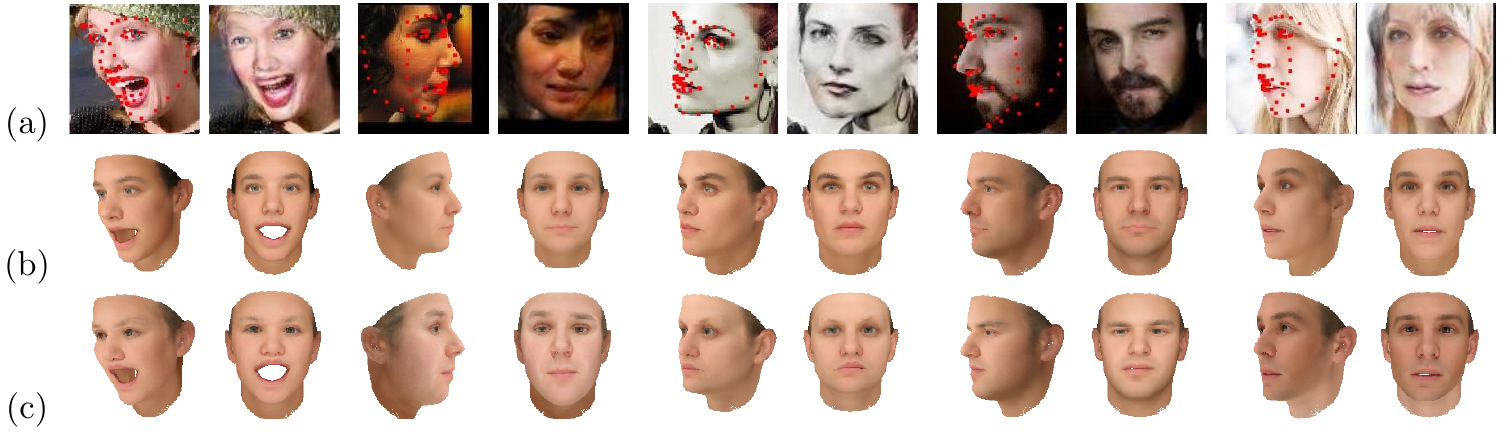}
\end{center}
\vspace{-6mm}
\caption{(a) our landmark localization and face frontalization results; (b) our 3DMM estimation; (c) ground truth from~\cite{zhu2016face}.}
\label{fig:3dmm}
\vspace{-5mm}
\end{figure*}

\subsection{Settings and Databases}
\vspace{-2mm}
We evaluate our proposed FF-GAN on a variety of tasks including face frontalization, 3D face reconstruction and face recognition. Frontalization and 3D reconstruction are evaluated qualitatively by comparing the visual quality of the generated images to ground truth. We also report quantitative results on sparse 2D landmark localization accuracy. Face recognition is evaluated quantitatively over several challenging databases. We pre-process the images by applying some state-of-the-art face detection~\cite{yangfancvpr2016} and face alignment~\cite{yu2016} algorithms and crop to $100\times100$ image size across all the databases. The face databases used for training and testing are introduced below.

{\bf{300W-LP}} consists of $122,450$ images that are augmented from 300W~\cite{300w2013} by the face profiling approach~\cite{ZhuXYCVPR15}, which is designed to generate images with yaw angles ranging from $-90^\circ$ to $90^\circ$. We use 300W-LP as our training set by forming image pairs of pose-variant and frontal-view images of the same identity. The estimated 3DMM coefficients provided with dataset are treated as ground truth to train module $R$.

{\bf{AFLW2000}} is constructed for 3D face alignment evaluation by the same face profiling method applied in 300W-LP. The database includes 3DMM coefficients and augmented $68$ landmarks for the first $2,000$ images in AFLW. We use this database to evaluate module $R$ for reconstruction. 

{\bf{Multi-PIE}} consists of $754,200$ images from $337$ subjects with large variations in pose, illumination and expression. We select a subset of $301,600$ images with $13$ poses, $20$ illuminations, neutral expression from all four sessions. The first $200$ subjects are used for training and the remaining $137$ subjects for testing, in the setting of \cite{tran2017}. We randomly choose one image for each subject with frontal pose and neutral illumination as gallery and all the rest as probe images.

{\bf{CASIA}} consists of $494, 414$ images of $10, 575$ subjects where the images of $26$ overlapping subjects with IJB-A are removed. It is a widely applied large-scale database for face recognition. We apply it to pre-train and finetune module $C$.

{\bf{LFW}} contains $13,233$ images collected from Internet. The verification set consists of $10$ folders, each with $300$ same-person pairs and $300$ different-person pairs. 
We evaluate face verification performance on frontalized images and compare with previous frontalization algorithms. 

{\bf{IJB-A}} includes $5,396$ images and $20,412$ video frames for $500$ subjects, which is a challenging with uncontrolled pose variations. Different from previous databases, IJB-A defines face template matching where each template contains a variant amount of images. It consists of $10$ folders, each of which being a different partition of the full set. We finetune model $C$ on the training set of each folder and evaluate on the testing set for face verification and identification.


\subsection{3D Reconstruction}
\vspace{-2mm}
FF-GAN borrows prior shape and appearance information from 3DMM to serve as the reference for frontalization. Though we do not specifically optimize for the reconstruction task, it is interesting to see whether our reconstructor is doing a fair job in the 3D reconstruction task.

Figure~\ref{fig:3dmm} (a) shows five examples on AFLW2000 for landmark localization and frontalization. Our method localizes the key points correctly and generates realistic frontal faces even for extreme profile inputs. We also quantitatively evaluated the landmark localization performance using the normalized mean square error. Our model $R$ achieves $6.01$ normalized mean square error, compared to $5.42$ for 3DDFA~\cite{zhu2016face} and $6.12$ for SDM~\cite{xiong2013}. Note that our method achieves competitive performance compared to 3DDFA and SDM, even though those methods are tuned specifically for the localization task. This indicates that our reconstruction module performs well in providing correct geometric information. 

Given the input images in (a), we compute the 3DMM coefficients with our model $R$ and generate the 3D geometry and texture using \eqref{eq:3d}, as shown in Figure~\ref{fig:3dmm} (b). We observe that our method effectively preserves shape and identity information in the estimated 3D models, which can even outperform the ground truth provided by 3DDFA. For example, the shape and texture estimations in the last example is more similar to the input while the ground truth clearly shows a male subject rather than a female. 
Given that the 3DMM coefficients cannot preserve local appearance, we obtain such high frequency information from the input image. Thus, the choice of fusing 3DMM coefficients with the original input is shown to be a reasonable one empirically.

\subsection{Face Recognition}
\vspace{-2mm}
One motivation for face frontalization is to see, whether the frontalized images bring in the correct identity information for the self-occluded missing part, and thus boost the performance in face recognition. To verify this, we evaluate our model $C$ on LFW~\cite{lfwdatabase}, MultiPIE~\cite{multipie}, and IJB-A~\cite{ijba} for verification and identification tasks. Features are extracted from model $C$ across all the experiments. Euclidean distance is used as the metric for face matching.

\begin{figure}
\begin{center}
\includegraphics[width=0.46\textwidth]{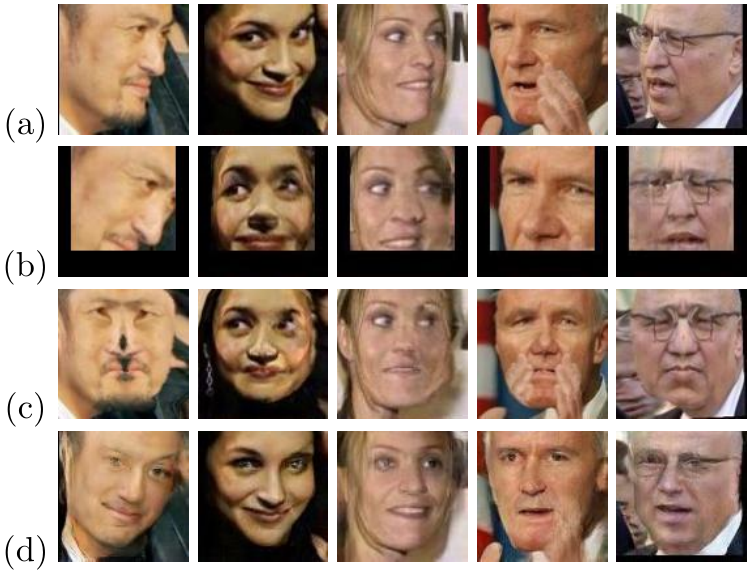}
\end{center}
\vspace{-6mm}
\caption{Face frontalization results on LFW. (a) Input; (b) LFW-3D~\cite{hassner2014}; (c) HPEN~\cite{ZhuXYCVPR15}; (d) FF-GAN (ours). }
\label{fig:lfw}
\vspace{-4mm}
\end{figure}

\noindent{\bf{Evaluation on LFW}}
We evaluate the face verification performance on our frontalized images of LFW, compared to previous face frontalization methods. LFW-FF-GAN represents our method to generate frontalized images, LFW-3D is from \cite{hassner2014}, and LFW-HPEN from \cite{zhu2016face}. Those collected databases are pre-processed in {\it the same way} as ours. As shown in Table \ref{tab:lfw}, our method achieves strong results compared to the state-of-the-art methods, which verifies that our frontalization technique better preserves identity information. Figure~\ref{fig:lfw} shows some visual examples. Compared to the state-of-the-art, our method can generate realistic and identity-preserving faces especially for large poses. 
The facial detail filling technique in~\cite{zhu2016face} relies on a symmetry assumption and may lead to inferior results ($1$st row, $4$th column). 
In contrast, we introduce a symmetry loss in the training process that generalizes to test images without the need to impose symmetry as a hard constraint for post-processing. 

\begin{table}[t]
\small
\begin{center}
\begin{tabular}{lcc}
\hline
Database & ACC($\%$) & AUC($\%$) \\ \hline
Ferrari et al.~\cite{ferrari2016} & - & $94.29$ \\ 
LFW-3D~\cite{hassner2014} & $93.62\pm1.17$ & $98.36\pm0.06$ \\
LFW-HPEN~\cite{ZhuXYCVPR15} & $96.25\pm0.76$ & $99.39\pm0.02$ \\ \hline
FF-GAN (syn.) & ${\bf96.42}\pm0.89$ & ${\bf99.45}\pm0.03$ \\ \hline
\end{tabular}
\end{center}
\vspace{-6mm}
\caption{\small Face verification accuracy (ACC) and area-under-curve (AUC) results on LFW.} 
\label{tab:lfw}
\vspace{-4mm}
\end{table}

\noindent{\bf{Evaluation on IJB-A}}
We further evaluate our algorithm on IJB-A database. Following prior work~\cite{tran2017}, we select a subset of well aligned images in each template for face matching. 
We define our distance metric as the original image pair distance plus the weighted generated image pair distance. The weights are the confidence score provided by our model $D$, i.e. $D(G({\bf{x}, \bf{p}}))$. Recall that model $D$ is trained for real or generated classification task, which reflects the quality of the generated images. Obviously, the poorer quality of the generated image, the less likely we take the generated image pair for distance metric.
With the fused metric distance, we expect the generated images to provide complimentary information to boost the recognition performance.

Table~\ref{tab:ijba} shows the verification and identification performance. On verification, our method achieves consistently better accuracy compared to the baseline methods. The gap is $6.46\%$ at FAR $0.01$ and $11.13\%$ at FAR $0.001$, which is a significant improvement. On identification, our fused metric also achieves consistently better result, $4.95\%$ improvement at Rank-$1$ and $1.66\%$ at Rank-$5$. As a challenging face database in the wild, large pose variations, complex background and the uncontrolled illumination prevent the compared methods to perform well. Closing one of those variation gaps would lead to large improvement, as evidenced by our face frontalization method in rectifying pose variation. 

\begin{table}[t]
\footnotesize
\begin{center}
\begin{tabular}{@{\hskip .5mm}l@{\hskip 1.5mm}c@{\hskip 1.5mm}c@{\hskip 1.5mm}c@{\hskip 1.5mm}c@{\hskip .5mm}}
\hline
Method $\downarrow$ & \multicolumn{2}{c}{Verification} & \multicolumn{2}{c}{Identification} \\ \hline
Metric ($\%$) $\to$ & @FAR=$0.01$ & @FAR=$0.001$ & @Rank-$1$ & @Rank-$5$ \\ \hline\hline
OpenBR~\cite{ijba} & $23.6\pm0.9$ & $10.4\pm1.4$ & $24.6\pm1.1$ & $37.5\pm0.8$ \\
GOTS~\cite{ijba} & $40.6\pm1.4$ & $19.8\pm0.8$ & $44.3\pm2.1$ & $59.5\pm2.0$ \\
Wang et al.~\cite{wang2016face} & $72.9\pm3.5$ & $51.0\pm6.1$ & $82.2\pm2.3$ & $93.1\pm1.4$ \\
PAM~\cite{MasiCVPR16} & $73.3\pm1.8$ & $55.2\pm3.2$ & $77.1\pm1.6$ & $88.7\pm0.9$ \\
DCNN~\cite{chen2016unconstrained} & $78.7\pm4.3$ & \textendash & $85.2\pm1.8$ & $93.7\pm1.0$ \\ 
DR-GAN~\cite{tran2017} & $77.4\pm2.7$ & $53.9\pm4.3$ & $85.5\pm1.5$ & $94.7\pm1.1$ \\\hline
FF-GAN (fuse) & ${\bf{85.2}}\pm1.0$ & ${\bf{66.3}}\pm3.3$ & ${\bf{90.2}}\pm0.6$ & ${\bf{95.4}}\pm0.5$ \\ \hline
\end{tabular}
\end{center}
\vspace{-6mm}
\caption{\small Performance comparison on IJB-A database. }
\label{tab:ijba}
\vspace{-4mm}
\end{table}

\noindent{\bf{Evaluation on Multi-PIE}}
Multi-PIE allows for a graded evaluation with respect to pose, illumination, and expression variations. Thus, it is an important database to validate the performance of our methods with respect to prior works on frontalization. The rank-$1$ identification rate is reported in Table~\ref{tab:multipie}. Note that previous works only consider poses within $60^\circ$, while our method can handle all pose variation including profile views at $90^{\circ}$. The results suggest that when pose variation is within $15^{\circ}$, which is near frontal, our method is competitive to state-of-the-art methods. But when the pose is $30^{\circ}$ or larger, our method demonstrates significant advantages over all the other methods. Although the recognition rate of the synthetic images performs worse than the original images, the fused results perform better than the original images, especially on large-pose face images.
\begin{table}[t]
\footnotesize
\label{tab:MTPIE_results} 
\begin{center}
\begin{tabular}{@{}l@{\hskip 1mm}c@{\hskip 1.3mm}c@{\hskip 1mm}c@{\hskip 1mm}c@{\hskip 1mm}c@{\hskip 1mm}c@{\hskip 1mm}c@{\hskip 1mm}c@{\hskip 1mm}c@{\hskip 1mm}c@{\hskip 1mm}c@{\hskip 1.3mm}c@{\hskip 1.3mm}c@{} }
\hline
& $0^o$ & $15^o$ & $30^o$ & $45^o$ & $60^o$ & $75^o$ & $90^o$ & avg$1$ & avg$2$\\ \hline
Zhu et al.~\cite{zhu2013deep} & $94.3$ & $90.7$ & $80.7$ & $64.1$ & $45.9$ & \textendash& \textendash &$72.9$ & \textendash \\
Zhu et al.~\cite{ZhuNIPS2014} & $95.7$ & $92.8$ & $83.7$ & $72.9$ & $60.1$ & \textendash & \textendash &$79.3$ & \textendash \\
Yim et al.~\cite{yim2015} & $\bf{99.5}$ & $\bf{95.0}$ & $88.5$ & $79.9$ & $61.9$ & \textendash & \textendash & $83.3$ & \textendash \\
DR-GAN~\cite{tran2017} & $97.0$ & $94.0$ & $90.1$ & $86.2$ & $83.2$ & \textendash & \textendash& $89.2$ & \textendash \\ \hline
FF-GAN (ori.) & $95.6$ & $\bf{95.0}$ & $\bf{92.6}$ & $\bf{89.8}$ & $84.3$ & $75.9$ & $58.2$ & $91.5$ & $84.5$ \\ 
FF-GAN (syn.) & $94.9$ & $86.1$ & $82.0$ & $77.0$ & $70.8$ & $62.1$ & $46.2$ & $82.1$ & $74.2$ \\ 
FF-GAN (fuse) & $95.7$ & $94.6$ & $92.5$ & $89.7$ & $\bf{85.2}$ & $\bf{77.2}$ & $\bf{61.2}$ & $\bf{91.6}$ & $\bf{85.2}$ \\ \hline
\end{tabular}
\end{center}
\vspace{-6mm}
\caption{\small Performance comparison on Multi-PIE database. Avg$1$ and avg$2$ are the average accuracy in $0$-$60^\circ$ and $0$-$90^\circ$ respectively.}
\vspace{-4mm}
\label{tab:multipie}
\end{table}

\begin{figure*}
\begin{center}
\includegraphics[width=.95\textwidth]{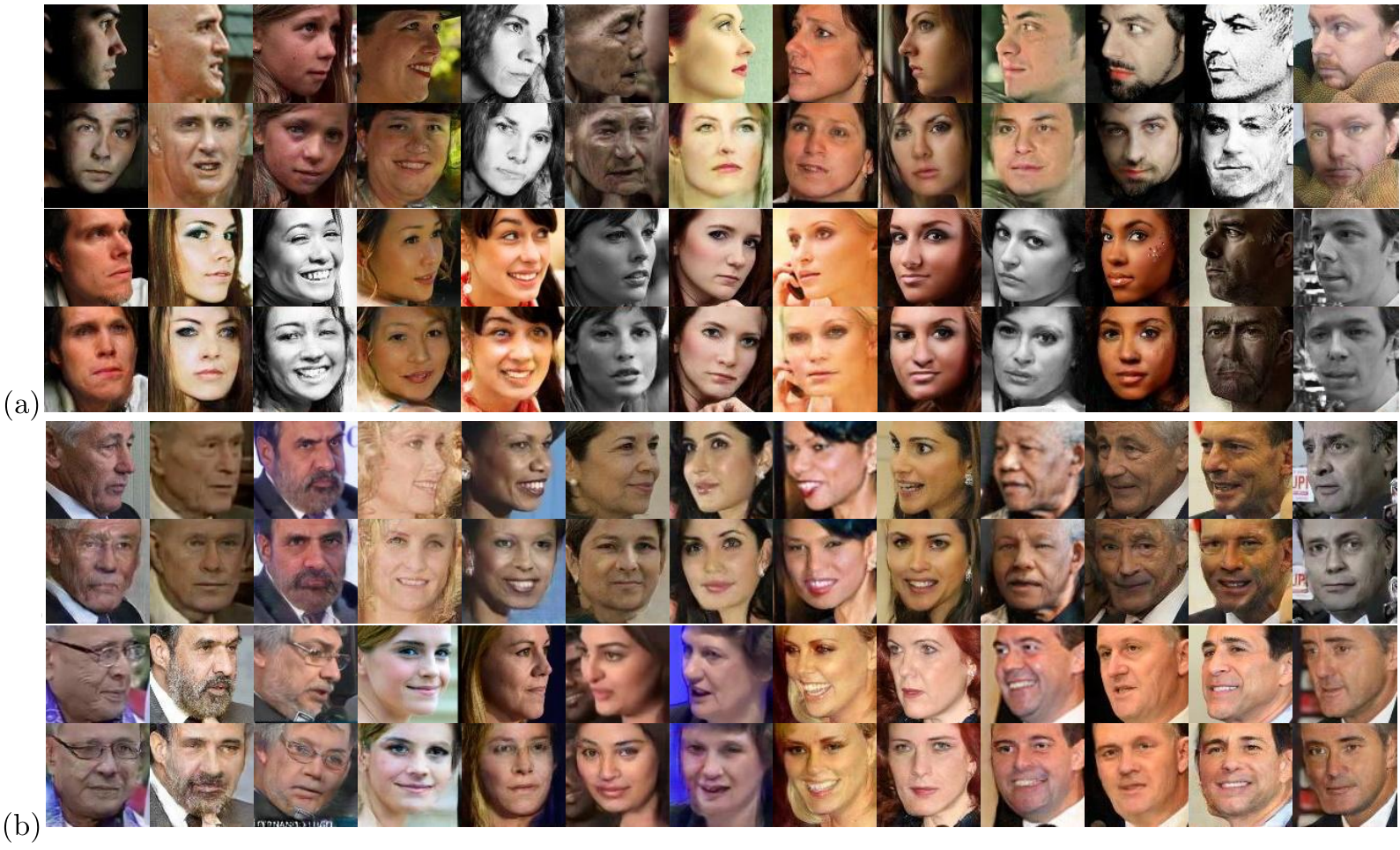}
\end{center}
\vspace{-6mm}
\caption{Visual results of our method on large-pose cases from different databases. (a) AFLW2000, (b) IJB-A.}
\label{fig:visual_all}
\vspace{-4mm}
\end{figure*}

\subsection{Qualitative Results}
\vspace{-2mm}
Figure~\ref{fig:visual_all} shows some visual results for unseen images on Multi-PIE, AFLW, and IJB-A.
The input images are of medium to large pose and under a large variation of race, age, expression, and lighting conditions. 
However, FF-GAN can still generate realistic and identity-preserved frontal faces.



\subsection{Ablation Study on Multi-PIE}
\vspace{-2mm}

\begin{figure}
\begin{center}
\includegraphics[width=0.46\textwidth]{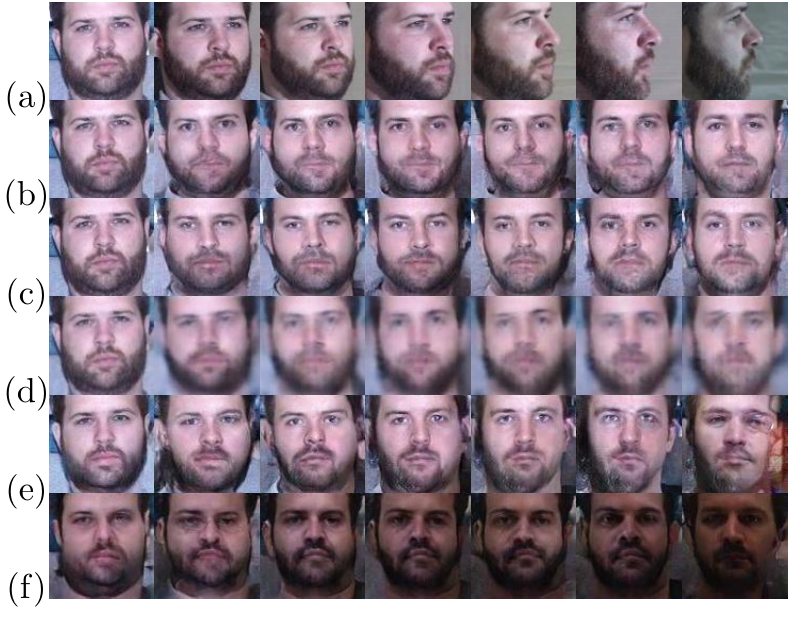}
\end{center}
\vspace{-6mm}
\caption{Ablation study results. (a) input images. (b) $\mathbb{M}$ (ours). (c) $\mathbb{M} \backslash \{C\}$. (d)$\mathbb{M} \backslash \{D\}$. (e) $\mathbb{M} \backslash \{R\}$. (f) $\mathbb{M} \backslash \{G_{id}\}$.}
\label{fig:ablation}
\vspace{-4mm}
\end{figure}

FF-GAN consists of four modules $\mathbb{M}=\{R,G,C,D\}$. Our generator $G$ is the key component for image synthesis, which cannot be removed. We train three partial variants by removing each of the remaining modules, which results in $\mathbb{M} \backslash \{C\}$, $\mathbb{M} \backslash \{D\}$, and $\mathbb{M} \backslash \{R\}$. Further, we train another three variants by removing each of the three loss functions (including $G_{id}$, $G_{tv}$, $G_{sym}$) applied on the generated images, resulting in $\mathbb{M} \backslash \{G_{id}\}$, $\mathbb{M} \backslash \{G_{tv}\}$, and $\mathbb{M} \backslash \{G_{sym}\}$. We keep the training process and all hyper-parameters the same and explore how the performances of those models differ. 

Figure \ref{fig:ablation} shows visual comparisons between the proposed framework and its incomplete variants. Our method is visually better than those variants, across all different poses, which suggests that each component in our model is essential for face frontalization. Without the recognizer $C$, it is hard to preserve identity especially on large poses. Without the discriminator $D$, the generated images are blurry without much high-frequency identity information. Without the reconstructor $R$, there are artifacts on the generated faces, which highlights the effectiveness of 3DMM in frontalization. Without the reconstruction loss $G_{id}$, the identity can be preserved to some extent, however the overall image quality is low and the lighting condition is not preserved. 

Table~\ref{tab:ablation} shows the quantitative results of the ablation study models by evaluating the recognition rate of the synthetic images generated from each model. Our FF-GAN with all modules and all loss functions performs the best among all other variants, which suggests the effectiveness of each part of our framework. For example, the performance drops dramatically without the recognition engine regularization. The 3DMM module also performs a significant role in face frontalization. 

\begin{table}[h]
\small
\begin{center}
\begin{tabular}{@{}l@{\hskip 2.2mm}c@{\hskip 2.2mm}c@{\hskip 2.2mm}c@{\hskip 2.2mm}c@{\hskip 2.2mm}c@{\hskip 2.2mm}c@{\hskip 2.2mm}c@{} }
\hline
removed module & $-$ & $C$ & $D$ & $R$ & $G_{id}$ & $G_{tv}$ & $G_{sym}$ \\ \hline
performance (syn.) & $\bf{74.2}$ & $59.2$ & $73.4$ & $68.5$ & $69.3$ & $72.9$ & $73.1$ \\ \hline
\end{tabular}
\end{center}
\vspace{-6mm}
\caption{\small Quantitative results of ablation study.}
\label{tab:ablation} 
\vspace{-4mm}
\end{table}


\vspace{-2mm}
\section{Conclusions}
\vspace{-2mm}
In this work, we propose a 3DMM-conditioned GAN framework to frontalize faces under all pose ranges including extreme profile views. 
To the best of our knowledge, this is the first work to expand pose ranges to $90^{\circ}$ in challenging large-scale in-the-wild databases.
The 3DMM reconstruction module provides an important shape and appearance prior to guide the generator to rotate faces. 
The recognition engine regularizes the generated image to preserve identity. 
We propose new losses and carefully design the training procedure to obtain high-quality frontalized images. 
Extensive experiments suggest that FF-GAN can boost face recognition performance and be applied for 3D face reconstruction. 
Large-pose face frontalization in the wild is a challenging and ill-posed problem; we believe this work has made convincing progress towards a viable solution.

{\small
\bibliographystyle{ieee}
\bibliography{egbib}
}

\end{document}